\documentclass[letterpaper]{article} 
\usepackage{aaai25}  
\usepackage{times}  
\usepackage{helvet}  
\usepackage{courier}  
\usepackage[hyphens]{url}  
\usepackage{graphicx} 
\usepackage{natbib}  
\usepackage{caption} 
\usepackage{algorithm}
\usepackage{algpseudocode}
\usepackage{booktabs}
\usepackage{amsmath}
\usepackage{multirow}
\usepackage{amssymb}  
\usepackage{bm}       

\usepackage{colortbl}
\usepackage[table]{xcolor}
\usepackage{booktabs} 
\usepackage{amsmath}

\usepackage{newfloat}
\usepackage{listings}
\DeclareCaptionStyle{ruled}{labelfont=normalfont,labelsep=colon,strut=off} 
\floatstyle{ruled}
\newfloat{listing}{tb}{lst}{}
\floatname{listing}{Listing}
\title{UniTT-Stereo: Unified Training of Transformer for Enhanced Stereo Matching}
\author{
    Soomin Kim\textsuperscript{\rm 1}, Hyesong Choi\textsuperscript{\rm 1}, Jihye Ahn\textsuperscript{\rm 1}, Dongbo Min\textsuperscript{\rm 1, *}
    }
\affiliations{

    \textsuperscript{\rm 1}Ewha W. University\\
    
    kim.soomin@ewha.ac.kr, hyesong@ewha.ac.kr, ajh0531@ewha.ac.kr, dbmin@ewha.ac.kr
}

\begin{document}

\maketitle

\begin{abstract}
Unlike other vision tasks where Transformer-based approaches are becoming increasingly common, stereo depth estimation is still dominated by convolution-based approaches. This is mainly due to the limited availability of real-world ground truth for stereo matching, which is a limiting factor in improving the performance of Transformer-based stereo approaches.
In this paper, we propose UniTT-Stereo, a method to maximize the potential of Transformer-based stereo architectures by unifying self-supervised learning used for pre-training with stereo matching framework based on supervised learning. 
To be specific, we explore the effectiveness of reconstructing features of masked portions in an input image and at the same time predicting corresponding points in another image from the perspective of locality inductive bias, which is crucial in training models with limited training data. Moreover, to address these challenging tasks of reconstruction-and-prediction, we present a new strategy to vary a masking ratio when training the stereo model with stereo-tailored losses. 
State-of-the-art performance of UniTT-Stereo is validated on various benchmarks such as ETH3D, KITTI 2012, and KITTI 2015 datasets. 
Lastly, to investigate the advantages of the proposed approach, we provide a frequency analysis of feature maps and the analysis of locality inductive bias based on attention maps. 


\end{abstract}

%

\section{Introduction}

Stereo matching remains fundamental for various computer vision applications, including autonomous driving, 3D reconstruction, and the recognition of objects~\cite{3d-application-1,3d-application-2}. The goal is to estimate a pixel-wise disparity map from two (or more) images capturing the same scene from distinct viewpoints, typically achieved by computing disparity from corresponding pixels.
The process of stereo matching is divided into two main parts: (1) feature matching and (2) disparity refinement. 
The key is to calculate the matching cost from two image pairs for feature matching and refine it accurately to obtain a reliable disparity map, considering challenges such as low-texture areas and occlusions.

While most approaches~\cite{3dcost-1, 3dcost-2, 4dcost-1, 4dcost-2} adopt convolutional neural networks (CNNs) for extracting stereo feature and aggregating cost volume, recent studies~\cite{stereotransformer, sttr, goat, CroCov2} have attempted to utilize the Transformer architecture, which is known to have superior representation capabilities and larger receptive fields compared to traditional CNNs.
It is reported that attention mechanisms within the Transformer framework can effectively replace the traditional cost volume approaches. This enables for the dense computation of correlation between two high-resolution features without being constrained by pre-defined disparity search range unlike cost volume approaches.

Nevertheless, the performance of Transformer-based stereo approaches is at best comparable or even inferior to that of convolution-based approaches, which means that  the Transformer architecture is not yet fully utilized in the context of stereo matching. 
To maximize the advantages of the Transformer while addressing its under-utilization in the stereo matching, the characteristics of the Transformer and stereo task need to be examined thoroughly. 
Recent research on the Transformer~\cite{Deit, data-efficient-1,data-efficient-3} suggests that it demands more training data for ensuring convergence due to the lack of inductive bias compared to CNNs, which benefit from structural characteristics like local receptive fields. 
In contrast, a deep-seated challenge of the stereo task is the limited availability of real-world ground truth, primarily due to the requirements of specialized equipment such as active range sensors (e.g., LiDAR), leading to increased cost and complexity of collecting large-scale labeled training data. 
Thus, it is crucial to resolve the inductive bias deficit and effectively use \textit{stereo} information from limited stereo training data when utilizing the Transformer in the stereo matching task.

\begin{figure*}[tb]
  \centering
  \includegraphics[width= 14.5cm]{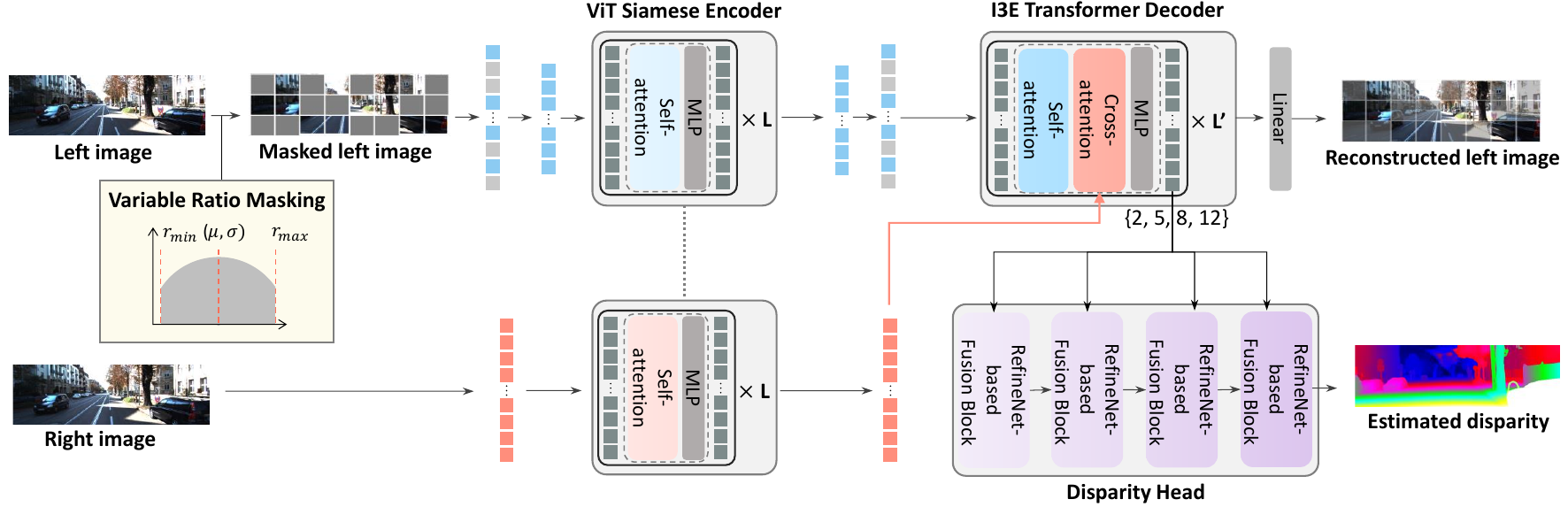}
  \caption{
  \textbf{Architecture Overview of UniTT-Stereo.}
  The visible tokens of masked left and original right images are fed into the Siamese ViT encoder for feature extraction, and then these image features are fed into the Inter Image Information Exchange (I3E) Transformer decoder based on cross-attention layers. The masked left image is reconstructed through a linear head while the disparity map is predicted by the RefineNet-based fusion module~\cite{refinenet}. Note that a masking ratio varies to ensure the model learns effectively across a range of information scales. The proposed reconstruction-and-prediction strategy introduces locality inductive bias in training the Transformer based stereo matching network, achieving competitive performance on various stereo benchmarks.
  }
  \label{fig:train-architecture}
\end{figure*}

In this context, we propose a novel approach, \textbf{UniTT-Stereo}, which stands for \textbf{Uni}fied \textbf{T}raining of \textbf{T}ransformer for enhanced stereo matching. Our model unifies self-supervised learning methods~\cite{SimMIM, MAE}, traditionally used for pre-training, with stereo matching framework based on supervised learning, enabling an effective learning tailored to Transformer based stereo matching. Our approach partially masks a left image and uses its remaining portions along with a right image to simultaneously reconstruct the original left image and estimate depth values for all pixels.
It is important to note that our experiments revealed that simply introducing the masking-and-reconstruction methodology does not necessarily improve performance. Therefore, our approach leverages a Variable Masking Ratio within the unified network, enabling the model to learn richer and more diverse information. Higher masking ratios facilitate the learning process during reconstruction, while lower masking ratios are advantageous for precise and detailed depth prediction. This balance ensures that our model can effectively capture both broad and fine-grained details, further enhancing its performance. 
We also adapt stereo-tailored losses to fully utilize the limited stereo information as mentioned earlier. We employ three synergistic loss functions at the final output level, RGB level, and feature level. Remarkably, we achieve improvements without the need for additional parameters or frameworks, relying solely on these well-designed approaches and stereo-tailored losses.

We also provided detailed analysis on the proposed approach at various aspects. First, the locality inductive bias of our approach using the reconstruction-and-prediction task is examined and compared with existing Transformer based stereo matching methods by computing attention distances (Fig.~\ref{fig:attention-distance}). This will be further validated by visualizing the attention maps of cross attention module between left and right features. Additionally, the ability to capture local patterns effectively may be related to exploiting high-frequency spatial information, which offers advantages in stereo tasks by improving the accuracy of depth estimation, particularly at object boundaries and fine details. To investigate how our method effectively amplifies high-frequency information, we perform Fourier analysis on the decoder's feature map used in the disparity head, in Fig.~\ref{fig:fourier}. Detailed explanations are provided in Analysis section.

We achieve state-of-the-art results on ETH3D~\cite{eth3d}, KITTI 2012~\cite{KITTI2012} and KITTI 2015 \cite{KITTI2015}
 datasets, demonstrating the effectiveness of our method. Our key contributions are as follows:

\begin{itemize}
    \item We examine the impact of the reconstruction-and-prediction approach on stereo depth estimation and propose a unified training approach based on these insights. 
    \item We enhanced performance by introducing a stereo-tailored combination of loss functions from multiple perspectives: feature, RGB, and disparity.
    \item Through extensive analyses and experiments, we validate that our model effectively leverages Transformer for stereo matching.

\end{itemize}

\section{Related works}
\subsection{Dense prediction with Transformer}
Fully convolutional networks~\cite{related-work-cnn-1,related-work-cnn-2} serve as the backbone for dense prediction, with various adaptations proposed over time. These architectures commonly depend on convolution and downsampling as fundamental components for acquiring multiscale representations, enabling the incorporation of a substantial contextual understanding. However, the low resolution in the deeper layer causes difficulty in dense prediction, so there have been many researches to maintain high resolution. 
Transformers~\cite{ViT, SwinTransformer, PyramidTransformer, chen2021pre, ranftl2021vision, lee2022knn, lee2023cross, hong2022cost}, based on the self-attention mechanism, demonstrate success with high-capacity architectures trained on extensive datasets. Since the Vision Transformer~\cite{ViT} adapts this mechanism to the image domain successfully but not in dense prediction, two main approaches have appeared. One is to design a specialized Transformer fitted to the dense prediction task~\cite{SwinTransformer, PyramidTransformer}, and the other is to use a plain Vision Transformer and the customized decoder for dense prediction. Dense Prediction Transformer~\cite{DPT} used the latter method and achieved state-of-the-art performance in 2021. We propose an approach to fully utilize transformer-based architecture for stereo depth estimation.

\subsection{Masked image modeling}
Masked image modeling (MIM) is a technique for self-supervised representation learning~\cite{grill2020bootstrap, chen2021exploring, ema1, ema2} using images that have masked parts. In this approach, some of the tokenized input sequence is replaced with trainable mask tokens, and the model is trained to predict the missing context based solely on the visible context. This approach, which does not require labels, is widely used for pre-training. SimMIM~\cite{SimMIM} and MAE~\cite{MAE} suggest that random masking with a higher mask ratio (e.g. 90\%) or size can perform well for self-supervised pretraining from image data. Recently, MTO~\cite{mto} has improved pre-training efficiency by optimizing masked tokens, while SBAM~\cite{sbam} has introduced a dynamic approach to the process with a saliency-based adaptive masking strategy that adjusts masking ratios according to the salience of the tokens. CroCo~\cite{CroCo} and CroCo v2~\cite{CroCov2} introduced a novel self-supervised pretraining approach exclusively designed for 3D tasks, reconstructing the masked image using the reference image. One of the advantages of Transformers is the abundance of these well-pretrained models available for use. Several studies~\cite{whatdo, mimocclusion, darksecret} have investigated the effects and what the model learns from MIM as a pretraining method compared to other approaches like contrastive learning. Meanwhile, through experimentation, we have identified how MIM can impact stereo tasks and the strategies to actively leverage MIM for the specific task of stereo depth estimation.
\subsection{Stereo depth estimation}

Stereo depth estimation is extensively used in fields such as autonomous driving~\cite{li2019stereo, chen2020dsgn}, robotics~\cite{wang2023application, nalpantidis2010stereo}, where accurate depth data is essential for navigation and object detection, and it is also increasingly employed as ground truth labels in monocular depth estimation tasks~\cite{tonioni2019unsupervised, choi2021adaptive}. Stereo depth estimation requires predicting a pixel-wise dense disparity map, capturing detailed and fine information, especially for boundary regions. In traditional deep stereo matching methods, the primary steps involve four components: feature extraction, cost volume creation, feature matching, and disparity regression. To enhance either accuracy or speed, researchers have proposed several strategies to improve these four components. 3D correlation cost volume~\cite{3dcost-1,3dcost-2} or 4D concatenation cost volume~\cite{4dcost-1,4dcost-2} can be constructed to measure the similarity between two views. Several studies~\cite{igev, raftstereo} have adopted iterative methods to construct disparity maps, and concurrent work~\cite{chen2024mocha} has also improved performance using this approach. Recent studies~\cite{stereo-attention-ref, stereotransformer, sttr, goat, CroCov2} have utilized cross-attention mechanisms to enable the exchange of information between different images instead of cost volume. We improve the performance by applying optimized approaches from an analytical perspective on the compatibility between Transformer architecture and stereo depth estimation task.

\section{Proposed Method}
\label{sec:methods}

We introduce MIM for effective learning by utilizing pairs of a masked left image and an unmasked right image. Unlike pre-training that focuses solely on reconstruction, our goal is to improve the performance of specific downstream tasks, and thus we consider the need for a more suitable masking method. To this end, we introduce variable ratio masking through a truncated normal distribution. After both image tokens pass through the Transformer encoder, we use cross-attention modules for inter-image information exchange. Finally, the model outputs a disparity map and a reconstructed image through respective heads. Our training process involves three losses: feature consistency loss, image reconstruction loss, and disparity loss. Fig.~\ref{fig:train-architecture} shows the overall architecture of the proposed method. Additionally, we provide an analysis of how our approach impacts the stereo task and enhances performance using attention distance, attention map, and Fourier Transform.

\subsection{Architecture}
Given left and right images \(I_l\) and \(I_r\), each of which captures the same scene from different viewpoints, both are split into \(N\) non-overlapping patches, denoted as \(p_l = \{p_l^1, ..., p_l^N\}\) and \(p_r = \{p_r^1, ..., p_r^N\}\). \(n= \left \lfloor rN \right \rfloor\) tokens are randomly masked only in the left image, where \(r\in [0,1]\) is a selected masking ratio. Siamese encoders deal with visible tokens from a left image and whole tokens from a right image independently. The encoder consists of 12 blocks with a dimension of 768 for ViT-Base and 24 blocks with a dimension of 1024 for ViT-Large. 

The left image tokens from the encoder are padded with masked tokens, resulting in \(F_l\) with \(N\) tokens, which is the same number as the tokens from the right image feature \(F_r\). The encoded left feature is then utilized by a decoder, conditional on the encoded right feature. The model constructs the query, key, and value in a self-attention block from the left token sequence in order to compute attention scores and identify relationships between tokens in the same sequence. In contrast, the model generates a cross-attention block by using the left token sequence to set up the query and the right token sequence to generate the key and value in order to find correspondences between the two images. It is composed of 12 blocks, each with a dimension of 768. 

For generating pixel-wise depth predictions, RefineNet-based fusion module is adapted to reshape and merge four features from different transformer decoder blocks. We utilize features from \(\{2, 5, 8, 12\}^{th}\) layers in the decoder. A linear head is used to get the reconstructed image output.

\subsection{Variable Ratio Masking}

Inspired from MIM pre-trained models, we leverage masked input to capture local and high-frequency patterns effectively. However, masking too much information can make it excessively difficult for the model to directly predict disparity maps, potentially hindering the model's ability to learn from raw RGB images. Conversely, when masking with a low ratio, there is no significant change in performance. To address this, we introduce a variable mask ratio, as shown in Algorithm~\ref{alg:masking}, to ensure the model learns effectively across a range of information scales. 


\begin{algorithm}[t]
\caption{Variable Ratio Masking}
\label{alg:masking}
\footnotesize
\raggedright
\textbf{Input:} \\
\hspace*{2.7em} Image tokens $I$ \\
\hspace*{2.7em} Mask size $m$ \\
\hspace*{2.7em} Image dimensions $h \times w$ \\
\hspace*{2.7em} Mask ratio range $[r_{min}, r_{max}]$ \\
\hspace*{2.7em} Mean $\mu$ and standard deviation $\sigma$ of mask ratio\\
\textbf{Output:} \\
\hspace*{2.7em} Masked image tokens

\begin{algorithmic}[1]
\State $N \gets \frac{h \times w}{m^2}$  \Comment{Total number of image tokens}
\State Initialize truncated normal distribution $TND(\mu, \sigma, r_{min}, r_{max})$
\State $r \gets \text{sample from } TND$ \Comment{Sample mask ratio}
\State $r \gets \text{round}(r,1)$ \Comment{Round ratio to one decimal place}
\State $n \gets \lfloor r \times N \rfloor$ \Comment{Calculate number of tokens to mask}
\State $M \gets \text{randomly select } n \text{ unique indices from } \{1, 2, \ldots, N\}$
\For{each index $i \in M$}
    \State $I[i] \gets \text{mask token}$ \Comment{Mask the selected tokens}
\EndFor
\State \Return $I$ \Comment{Return masked image tokens}
\end{algorithmic}
\end{algorithm}

We use random masking with mask size 16, similar to SimMIM or MAE~\cite{SimMIM, MAE} with variable ratio. The masking ratio is determined from a truncated normal distribution with specified upper and lower bounds, which is generated by the given mean and standard deviation. A new masking ratio is then randomly sampled from this distribution for each batch and it is rounded to one decimal place. For example, 0.32 is rounded to 0.3, resulting in 30\% masking. We confirmed the effectiveness of variable ratio masking through experiments in various settings. By default, we use a truncated normal probability distribution with a lower bound of 0.0, an upper bound of 0.5, a mean of 0.25, and a standard deviation of 1.0.


\subsection{Losses}

\subsubsection{Feature Consistency Loss} To the output feature maps from each encoder, we introduce the consistency loss which aims to enhance the alignment between two corresponding features from a stereo pair. This makes the model improve matching performance at a feature level. This is achieved by warping the feature from the right image to the feature from the left image, using the ground truth disparity information. \(\tilde{F_l}\) means reconstructed left feature from the right feature with disparity. 
\begin{equation}
  L_{consist}=\sum_{i}|F_{l,i}-\tilde{F_{l,i}}|
  \label{eq:consistency-loss}
\end{equation}

\subsubsection{Disparity Loss} The disparity loss is common and plain, but the most powerful matching loss which can be supervised by ground truth. 
We minimize negative log-likelihood with a Laplacian distribution to train the proposed model, following \cite{CroCov2}: 
\begin{equation}
   L_{disp}=\sum_{i}[\frac{|D _i-\bar{D_i}|}{s_i}-2\log s_i]
  \label{eq:disparity-loss}
\end{equation}
where \(D_i\) and \(\bar{D_i}\) are an estimated disparity and the ground truth disparity at pixel \(i\), respectively. The scale parameter \(s_i\) is also outputted from a model. It can be understood as an uncertainty score for predictions. 

\subsubsection{Image Reconstruction Loss} The reconstruction loss evaluates reconstruction accuracy by the Mean Squared Error (MSE) only for masked patches \(p_l\setminus \tilde{p_l}\) where \(p_l\) denotes a set of patches from the first image, \(\tilde{p_l}\) is a set of visible patches from the first image, and \(\hat{p_l}\) is the reconstructed first image. 
Notably, the left image undergoes reconstruction through image completion, utilizing corresponding information from the right image. Consequently, this loss can be considered as a form of matching loss from an RGB perspective.
\begin{equation}
  L_{recon}=\frac{1}{|p_l\setminus \tilde{p_l}|}\sum_{p_{l,i}\in p_l\setminus \tilde{p_l}} \left\| \hat{p_{l,i}}-p_{l,i}\right\|^2
  \label{eq:reconstruction-loss}
\end{equation}

\begin{figure}[tb]
  \centering
  \includegraphics[width=0.85\columnwidth]{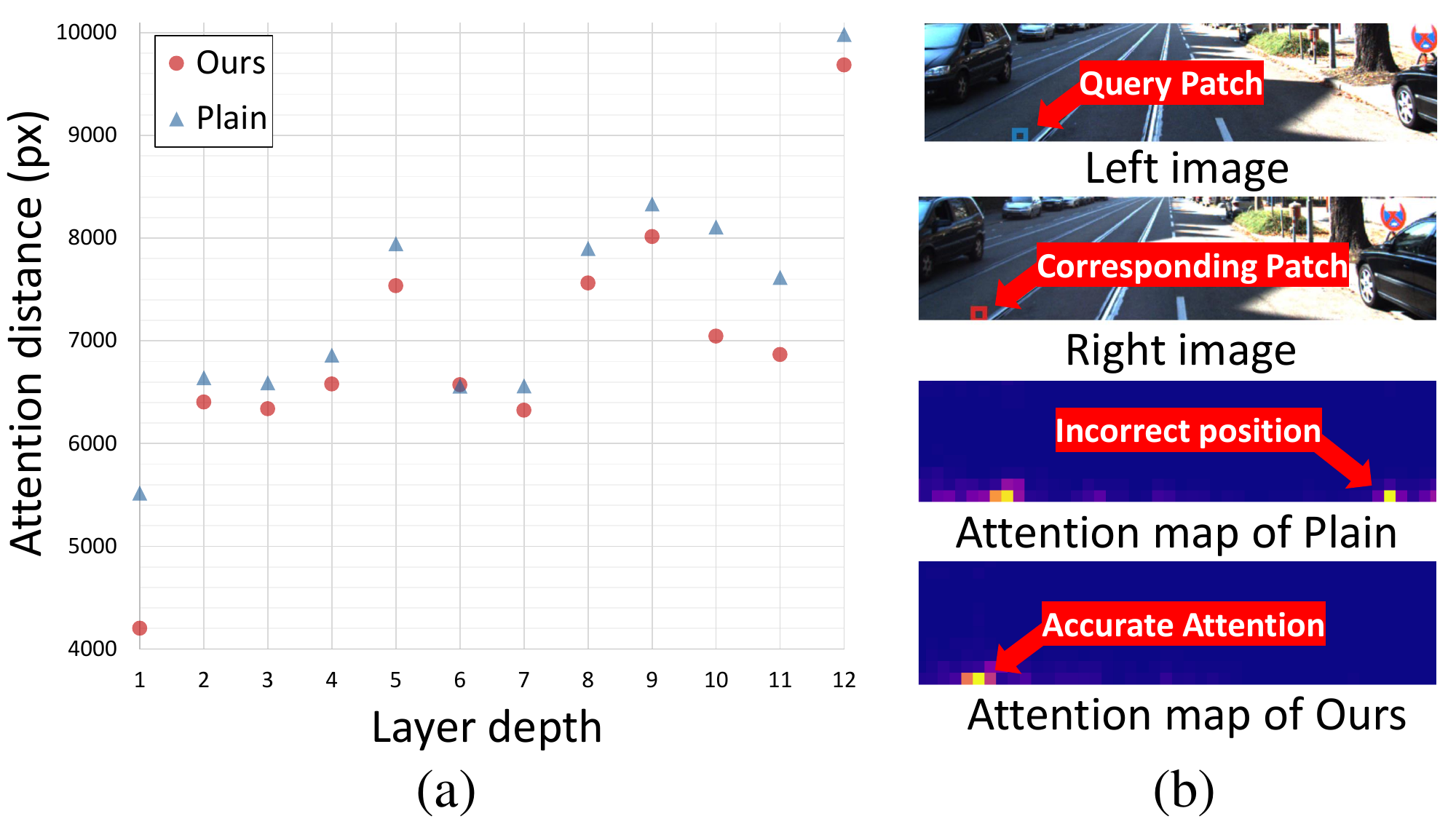}
  \caption{ \textbf{Attention Analysis}: \textbf{(a)} Attention distance plot; \textit{Plain} refers to the method where the same architecture is used with disparity loss alone for supervised learning, without incorporating our masking approach. \textit{Ours} refers to the case where our Unified Training method is applied. \textbf{(b)} Attention map visualization; Brighter colors indicate higher attention scores.
   }
  \label{fig:attention-distance}
\end{figure}

\begin{figure}[tb]
  \centering
  \includegraphics[width=0.85\columnwidth]{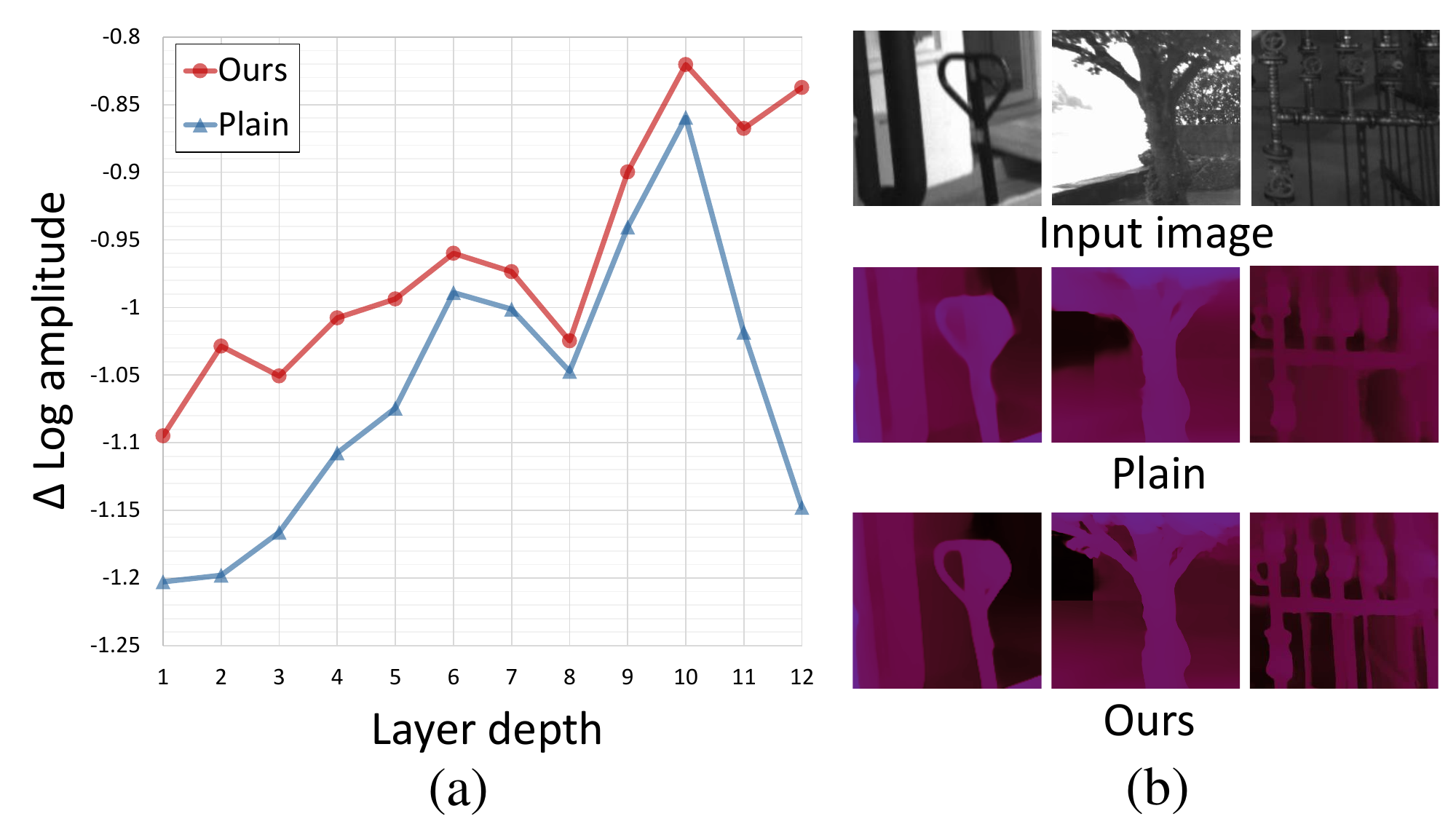}
  \caption{
  \textbf{Fourier Analysis}:
  \textbf{(a)} Fourier analysis of the feature maps of the decoder; The ratio of high-frequency components to low-frequency components is reported using the log amplitude metric. The log amplitude represents the difference in log amplitude between \(f = \pi\) (the highest frequency) and \(f = 0\) (the lowest frequency). This indicates how much the high-frequency components stand out compared to the low-frequency components. \textbf{(b)} The results from ETH3D test data; By amplifying and utilizing high-frequency information in the process of generating disparity maps, the resulting maps tend to have sharper boundaries and more fine-grained details. 
  }
  \label{fig:fourier}
\end{figure}

\begin{figure}[tb]
  \centering
  \includegraphics[width= 5.6cm]{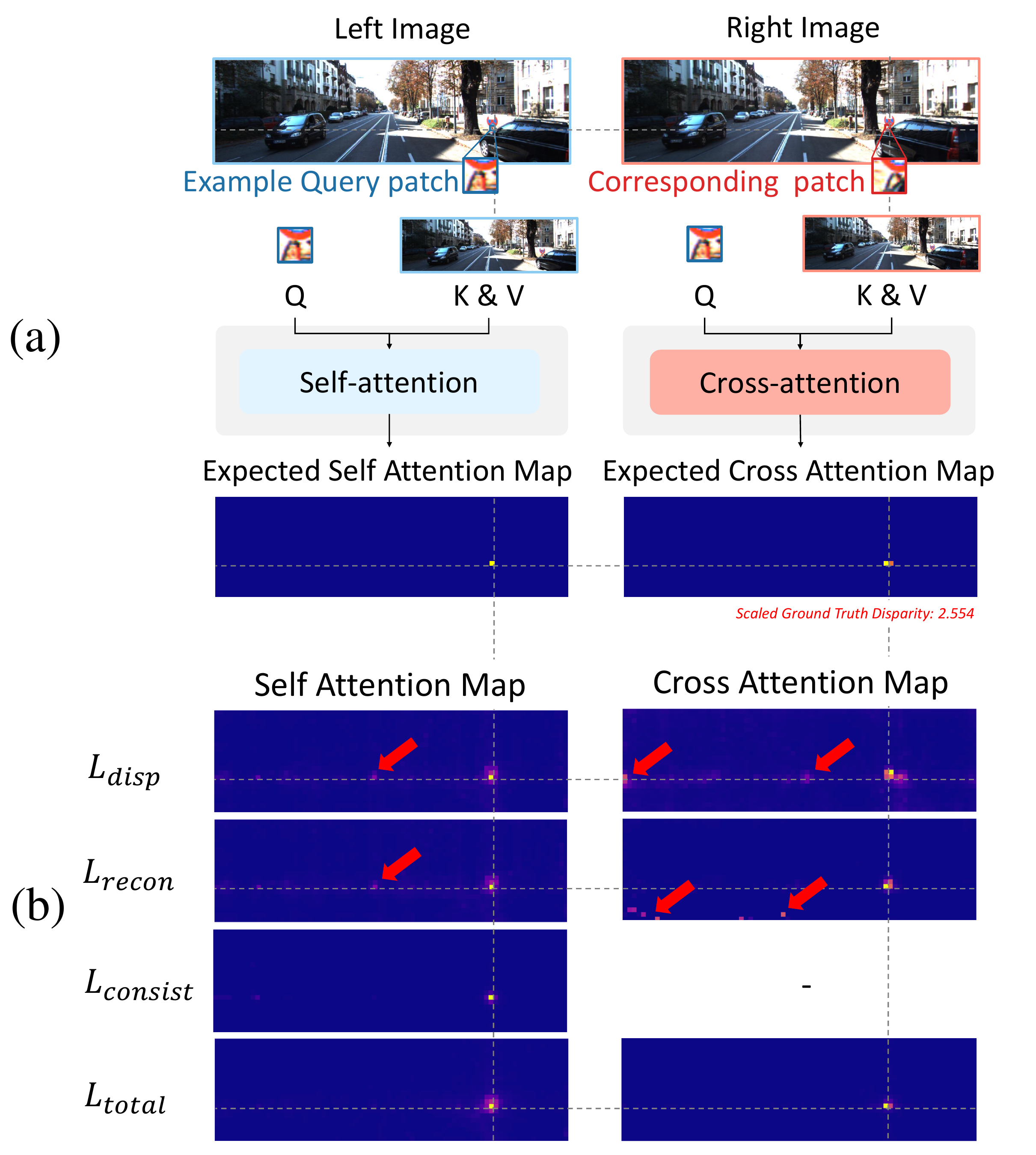}
  \caption{
  \textbf{Attention Map by Varying Losses}:
  \textbf{(a)} Example process of attention map visualization; Example query, key, and value are fed into the self-attention layer in the encoder or the cross-attention layer in the decoder. An expected attention map is created using the ground truth disparity value of the query to determine the location of the corresponding patch. \textbf{(b)} Attention map visualization when the model is trained solely using each individual loss; Brighter colors indicate higher attention scores. Since the consistency loss applies to features processed by the encoder and does not directly influence the decoder, we do not visualize the cross-attention map in the case where the model was trained using only the consistency loss.
  }
  \label{fig:loss-attention}
\end{figure}

\subsubsection{Total Loss} We supervise the model with a linear combination of three synergistic losses which are disparity loss from final output, reconstruction loss from reconstructed image, and consistency loss from feature map as \(\ L_{total}=\lambda*L_{disp}+L_{recon}+L_{consist}\)
\noindent where $\lambda$ is set to 3. Since our objective is to attain improved performance on stereo depth estimation, the disparity loss takes on more weight than other losses. 


Moreover, to mitigate the potential learning of erroneous information during the early stages of training due to masked images, we calculate an uncertainty score using the reconstruction loss, allowing us to assess how effectively the model has adjusted to masked images. We used a fixed value \(\tau =0.4\) as a threshold for generalized reconstruction error \(\phi (L_{recon}) = \text{tanh}(L_{recon})\) to decide whether to impose the disparity loss or not. 
As estimated disparity from unsteady reconstructed feature makes the model unstable, if \(\phi>\tau\), only \(L_{recon} + L_{consist}\) is used.

\subsection{Analysis}
\label{sec:analysis}

{\bf Locality Inductive Bias}: Fig.~\ref{fig:attention-distance} (a) illustrates the attention distances calculated from the attention scores obtained after passing the entire KITTI 2015~\cite{KITTI2015} test dataset through the cross attention module. Each point represents the average attention distance across 12 heads for each layer. Our method encourages the model to focus on these local patterns to reconstruct the masked parts using \emph{locality inductive bias} via MIM. This harmonizes well with Transformers, which adept at learning global information. To further investigate the effect of the locality inductive bias in our method, we visualize the attention map in Figure.~\ref{fig:attention-distance} (b). It visualizes cross attention maps for an example query patch from an left image, divided into a set of \(16\times16\) patches, in KITTI 2015 training dataset. 
Ideally, the attention score should be highest at the location of the corresponding patch in the left image, as identified by the ground truth disparity. Our approach, which tends to focus more on local information, helps prevent incorrect attention values, when compared with the plain method that has a higher risk of occasionally concentrating attention on patches located far away, as shown in the right part of the example map.

\vspace{3pt}

\noindent {\bf Fourier Analysis}: In Fig.~\ref{fig:fourier} (a), we additionally conducted Fourier analysis on feature maps of blocks in the decoder of our model in Fig.~\ref{fig:train-architecture}, following~\cite{fourier}. This experiment conducted on ETH3D dataset~\cite{eth3d} reveals that our approach tends to generate feature maps with a higher proportion of high-frequency components and indicates that the model is capturing detailed and sharp features, which can be advantageous for tasks like stereo depth estimation.
High frequency components of the decoder features used in the depth estimation head enable for more precise and detailed representations of object edges and fine structures in the disparity map. 
As illustrated in Fig.~\ref{fig:fourier} (b), by effectively utilizing high-frequency information, the proposed method achieved sharper boundaries on the disparity maps.

\vspace{3pt}

\noindent {\bf Attention Map by Varying Losses}: To demonstrate the synergy when the losses are used together, Fig.~\ref{fig:loss-attention} visualizes the attention maps when the model is separately trained using each loss. Masking was applied only when the model was trained with the reconstruction loss or the total loss. We averaged the attention scores from every head and self-attention layer in the left encoder for self-attention visualization, and applied the same approach using cross-attention layer in the decoder to cross-attention visualization. 
In the self-attention map, the score should be highest at the location of the query patch, whereas the score should peak at the location of the corresponding patch in the cross-attention map.
As shown in Fig.~\ref{fig:loss-attention} (b), when trained with each loss individually, incorrect attention values appear at various locations in each case. Even the disparity loss case, which is supervised with ground truth, has its limitations. However, when these losses are combined (\(L_{total}\)), they can correct each other's errors and emphasize common attention patterns, working synergistically to guide the model towards more accurate attention placement. This mutual reinforcement allows the model to learn more effectively, facilitating improved stereo matching performance, as also validated in the ablation study of Table~\ref{tab:ablation_checkmarks}.


\section{Experiments}

\begin{figure*}[tb]
  \centering
  \includegraphics[width= 15.8cm]{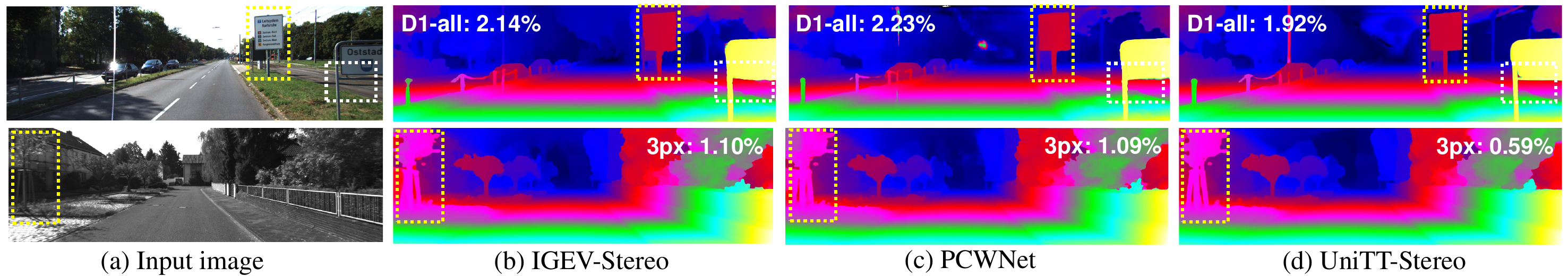}
  \caption{
  \textbf{Qualitative comparison on KITTI 2015 and 2012:}  The first row shows the result on KITTI 2015. UniTT-Stereo outputs clearer boundaries for objects compared to other models. The second row shows the result on KITTI 2012. Our model produces an accurate and sharp disparity map even in low texture areas with blurring.
  }
  \label{fig:kitti}
\end{figure*}

\begin{figure}[tb]
  \centering
  \includegraphics[width= 8.5cm]{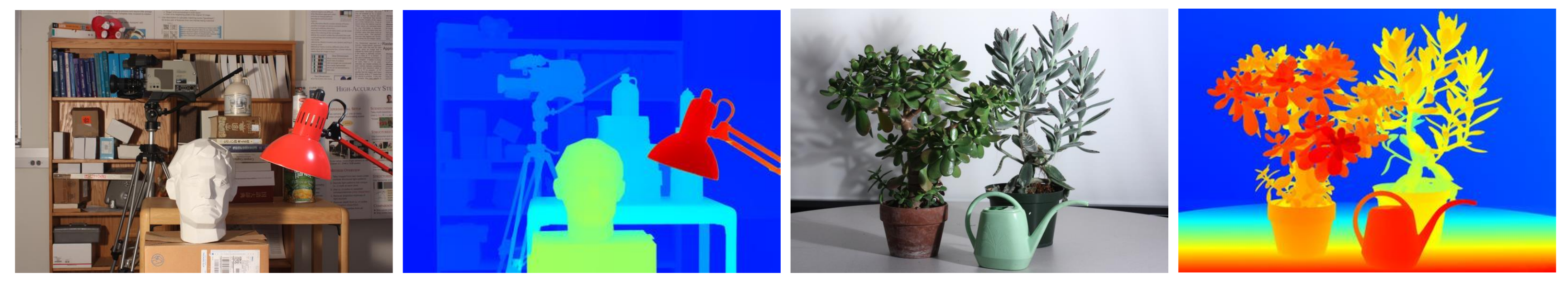}
  \caption{
  \textbf{Qualitative results on Middlebury evaluaton.} Our model demonstrates strong performance on sequences that require detailed depth prediction.
  }
  \label{fig:middlebury}
\end{figure}

\subsubsection{Implementation Details}
We train our UniTT-Stereo for 32 epochs using batches of 6 pairs initializing the encoder and decoder from the pre-trained weights by CroCov2~\cite{CroCov2}. For optimization, we employ the AdamW optimizer~\cite{adam} with a weight decay of 0.05. The learning rate of \(3\times10^{-5}\) follows a cosine schedule with a single warm-up epoch. We utilize SceneFlow~\cite{SceneFlow}, CREStereo~\cite{crestereo}, ETH3D~\cite{eth3d}, Booster~\cite{booster}, Middlebury (2005,
2006, 2014, 2021 and v3)~\cite{middlebury} with crop size of 704 × 352 to train UniTT-Stereo.
Afterward, we trained our model on KITTI 2012~\cite{KITTI2012} and KITTI 2015~\cite{KITTI2015} with crop size of 1216 × 352 for 100 epochs using effective batches of 6 pairs. We use a learning rate of \(3\times10^{-5}\). For inference, we use a tiling-based strategy in which we sample overlapping tiles with the same size as the training crops, following~\cite{CroCov2}. 

\subsection{Stereo Depth Estimation Performance}
We evaluate UniTT-Stereo on representative stereo datasets with their metrics and compare with the published state-of-the-art methods. 

\subsubsection{ETH3D}
UniTT-Stereo sets a new state-of-the-art on ETH3D. Table~\ref{tab:stereo-depth-eth3d} compares UniTT-Stereo with HITNet~\cite{hitnet}, RAFT-Stereo~\cite{raftstereo}, GMStereo~\cite{stereo-attention-ref}, IGEV-Stereo~\cite{igevstereo}, CREStereo~\cite{crestereo}, and CroCo-Stereo~\cite{CroCov2}. 

\setlength{\tabcolsep}{2.5pt}
\renewcommand{\arraystretch}{0.85} 
\begin{table}[t]
\setlength{\tabcolsep}{2\tabcolsep}
  \centering
    \fontsize{8pt}{10.5pt}\selectfont
    \caption{\textbf{Evaluation on ETH3D leaderboard}: Models were evaluated with the percentage of pixels with errors over 1px (bad@1.0), over 2px (bad@2.0), and the average error over non-occluded (noc) or all pixels.}
  \begin{tabular}{lcccccc}
    \toprule
        \multirow{2}{*}{Method}  & \multicolumn{2}{c}{bad@1.0\((\)\%\()\)$\downarrow$} & \multicolumn{2}{c}{bad@2.0\((\)\%\()\)$\downarrow$} & \multicolumn{2}{c}{avg err\((\)px\()\)$\downarrow$} \\
    &  noc & all & noc & all & noc & all\\
    \midrule
    HITNet & 2.79 & 3.11 & 0.80&1.01 & 0.20 & 0.22 \\
    RAFT-Stereo &  2.44 & 2.60 & 0.44&0.56 & 0.18 & 0.19 \\
    GMStereo & 1.83 & 2.07 & 0.25&0.34 & 0.19 & 0.21 \\
    IGEV-Stereo &  1.12 & 1.51 & \underline{0.21}&0.54 & \underline{0.14} & 0.20 \\
    CREStereo & \underline{0.98} & \underline{1.09} & 0.22& \underline{0.29} & \textbf{0.13} & \textbf{0.14} \\
    CroCo-Stereo & 0.99 & 1.14 & 0.39&0.50 & \underline{0.14} & \underline{0.15} \\
    \rowcolor{lightgray}
    \textbf{UniTT-Stereo} & \textbf{0.83} & \textbf{0.94} & \textbf{0.16} & \textbf{0.23} & \underline{0.14} & \underline{0.15}\\
    \bottomrule
  \end{tabular}
  \label{tab:stereo-depth-eth3d}
\end{table}


\subsubsection{KITTI 2012 \& 2015}

We also achieve state-of-the-art performance compared to other published methods, aside from concurrent work, on both KITTI 2012 and 2015. Table~\ref{tab:stereo-depth-kitti15} compares UniTT-Stereo with HITNet~\cite{hitnet}, PCWNet~\cite{pcwnet}, ACVNet~\cite{acvnet}, LEAStereo~\cite{leastereo}, CREStereo~\cite{crestereo}, IGEV-Stereo~\cite{igevstereo}, and CroCo-Stereo~\cite{CroCov2}. The qualitative comparison is shown in Fig.~\ref{fig:kitti}. 

\setlength{\tabcolsep}{2.5pt}
\begin{table}[t]
\setlength{\tabcolsep}{1\tabcolsep}
  \centering
  \fontsize{8pt}{10.5pt}\selectfont
    \caption{\textbf{Evaluation on KITTI 2015 and 2012 leaderboard}: For KITTI 2015, we evaluated along with the percentage of outliers for background (D1-bg), foreground (D1-fg), and all pixels combined (D1-all). For KITTI 2012, we provide the outlier ratio over $n$ pixel across all areas.}
  \begin{tabular}{lccccccc}
    \toprule
    \multirow{2}{*}{Method} & \multicolumn{3}{c}{KITTI 2015} & \multicolumn{3}{c}{KITTI 2012}\\
    & D1-bg$\downarrow$ & D1-fg$\downarrow$ & D1-all$\downarrow$ & 2px$\downarrow$ & 3px$\downarrow$ & 4px$\downarrow$ & 5px$\downarrow$ \\
    \midrule
    HITNet & 1.74 & 3.20 &  1.98 &2.65&1.89&1.53&1.29\\
    PCWNet & \underline{1.37} & 3.16 & 1.67 &2.18&\underline{1.37}&\underline{1.01}&\underline{0.81}\\
    ACVNet & \underline{1.37} & 3.07 &  1.65 &2.34&1.47&1.12&0.91\\
    LEAStereo & 1.40 & 2.91 &  1.65 &2.39&1.45&1.08&0.88\\
    CREStereo & 1.45 & 2.86 &  1.69 &2.18&1.46&1.14&0.95\\
    IGEV-Stereo & 1.38 & 2.67 & \underline{1.59} &\underline{2.17}&1.44&1.12&0.94\\
    CroCo-Stereo & 1.38 & \underline{2.65} &  \underline{1.59} &$-$&$-$&$-$&$-$\\
    \rowcolor{lightgray}
    \textbf{UniTT-Stereo} & \textbf{1.27} & \textbf{2.45} & \textbf{1.47}&\textbf{2.02}&\textbf{1.25}&\textbf{0.92}&\textbf{0.73}\\
    \bottomrule
  \end{tabular}
  \label{tab:stereo-depth-kitti15}
\end{table}

\subsubsection{Middlebury}

We also conducted performance evaluations on the Middlebury evaluation dataset. Table~\ref{tab:middlebury} compares UniTT-Stereo with LeaStereo~\cite{leastereo}, HITNet~\cite{hitnet}, RAFT-Stereo~\cite{raftstereo}, CREStereo~\cite{crestereo}, GMStereo~\cite{stereo-attention-ref}, and CroCo-Stereo~\cite{CroCov2}. As shown in Fig.~\ref{fig:middlebury}, our method delivered high performance on data that requires precise estimation.

\subsubsection{Limitations}
While our method achieved comparable results to other methods in Middlebury, there was a limitation in handling certain large maximum disparities leading to bad performance on several sequences. This is likely due to the constraints of tiling-based inference, which can restrict the ability to capture long-range correspondences. To address this, it may be necessary to use larger tile sizes or increase the overlap ratio.

\setlength{\tabcolsep}{5pt}
\renewcommand{\arraystretch}{0.8} 
\begin{table*}[t]
\centering
\scriptsize
\caption{\textbf{Evaluation on Middlebury leaderboard}: Models were evaluated with the average error over all pixels. UniTT-Stereo achieved comparable results overall. Through detailed and sharp predictions, our model ranked first on sequences where this precision is necessary. However, it showed limitations on sequences with large maximum disparity.}
\begin{tabular}{lcccccccccccccccc}
\toprule
  Method& Austr & AustrP & Bicyc2 & Class & ClassE & Compu &  Crusa & CrusaP & Djemb & DjembL &  Hoops & Livgrm & Nkuba & Plants &Stairs &   avg$\downarrow$ \\ 
\midrule
LEAStereo & 2.81 & 2.52 & 1.83 & 2.46 & 2.75 & 3.81 & 2.91 & 3.09 & 1.07 & 1.67 & 5.34 & 2.59 & 3.09 &  5.13 & 2.79 & 2.89 \\ 
HITNet& 3.61 & 3.27 & 1.43 &  2.43 & 3.20 & 1.87 & 4.67 & 4.74 & 0.90 & 9.12 & 4.45 & 2.37 & 3.45 & 4.07 & 3.38 & 3.29 \\ 
RAFT-Stereo& 2.64 & 2.22 & 0.90 &  \textbf{1.46} & 2.44 & \underline{1.13} & 4.58 & 6.00 &\textbf{0.63} & 1.22 & 3.54 & 3.13 & 4.36  & 3.55 & \underline{1.89} & 2.71 \\
CREStereo& 2.63 & 2.53 & 1.38 & \underline{1.92} & \underline{2.31} & \textbf{1.06} & \underline{1.78} & \underline{1.83} & \underline{0.64} & \textbf{1.11} & \underline{3.22} & \textbf{1.42} & 2.51 & 5.31 & 2.40 & \underline{2.10} \\ 
GMStereo & 2.26 & 2.23 & 1.34 & 2.19 & \textbf{2.08} & 1.32 &  \textbf{1.71} & \textbf{1.75} & 1.01 & 1.62 & \textbf{3.19} & \underline{1.84} & 2.10 & 2.49 & 2.18 & \textbf{1.89} \\
CroCo-Stereo & \underline{1.87} & \underline{1.83} & \underline{0.84} & 3.99 & 4.61 & 1.45 & 2.48 & 2.81 & 0.69 & \underline{1.19} & 8.31 & 2.40 & \underline{1.96} & \underline{2.28} & \textbf{1.44} & 2.36 \\ 
\rowcolor{lightgray}
\textbf{UniTT-Stereo}	& \textbf{1.51}	& \textbf{1.43}	& \textbf{0.74} & 4.97 & 5.75 & 1.97 & 1.98 & 2.35 & 0.66 & \textbf{1.11} & 8.60 & 2.17 & \textbf{1.72} & \textbf{1.82} & \textbf{1.44} & 2.32\\
\bottomrule
\label{tab:middlebury}
\end{tabular}
\end{table*}

\subsection{Zero-shot Generalization}
Generalizing from synthetic to real data is crucial due to the challenge of gathering real-world datasets. The result suggests that our approach help the model learn invariant features across different domains. Table~\ref{tab:DG} compares UniTT-Stereo with GANet~\cite{ganet}, RAFT-Stereo~\cite{raftstereo}, and DSMNet~\cite{dsmnet}. 

\setlength{\tabcolsep}{6pt}
\renewcommand{\arraystretch}{0.85} 
\begin{table}[t]
\caption{\textbf{Zero-shot generalization peformance}: In this evaluation, all methods including UniTT-Stereo are only trained on SceneFlow and tested on KITTI 2012, 2015, Middlebury (Quarter resolution), and ETH3D training set.}
\fontsize{8pt}{10.5pt}\selectfont
  \centering
  \begin{tabular}{lcccc}
    \toprule
    \multirow{2}{*}{Method}  & KITTI 12 & KITTI 15 & Middlebury (Q) & ETH3D\\
    &  $>$3px & $>$3px & $>$2px & $>$1px \\
    \midrule
    GANet & 10.1 & 11.7 & 11.2 & 14.1\\
    
    DSMNet & 6.2 & 6.5 & \textbf{8.1} & 6.2 \\
    CREStereo & \underline{4.7} & \textbf{5.2}& $-$ & \underline{4.4}\\
    RAFT-Stereo & \underline{4.7} & 5.5 & 9.4 & \textbf{3.3} \\
    \rowcolor{lightgray}
    \textbf{UniTT-Stereo} & \textbf{4.6} & \underline{5.4} & \underline{8.3} & 4.5 \\
    
    \bottomrule
  \end{tabular}
  \label{tab:DG}
\end{table}

\begin{figure}[t]
\centering
\begin{minipage}{0.65\columnwidth} 
    \centering
    \includegraphics[width=0.75\columnwidth]{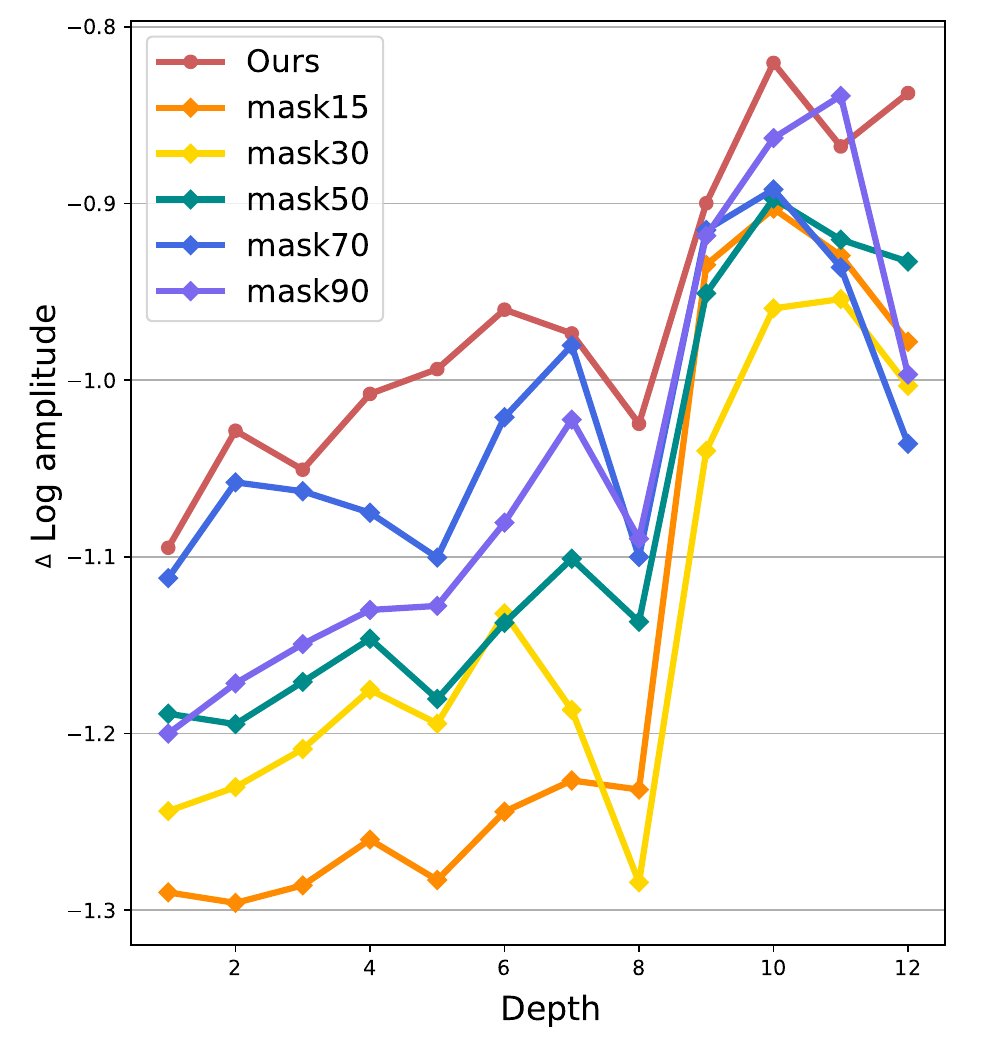}
\end{minipage}%
\begin{minipage}{0.35\textwidth} 
    \caption{ \\ \textbf{Fourier analysis \\according to the \\masking ratio}:\\
    \textit{Ours} refers to the \\ default settings \\mentioned in \\the Methods section.
  }
  \label{fig:fourier-mask-ratio}
\end{minipage}
\end{figure}

\subsection{Ablation Study}

\subsubsection{Key Components} We conducted an ablation study on the effectiveness of each key component, including masking and three loss functions. As listed in Table~\ref{tab:ablation_checkmarks}, introducing each additional loss function led to performance improvement in a sequential order. Optimal performance was achieved when employing every key component.

\subsubsection{Masking Ratio} We also evaluate the effectiveness of the variable ratio masking. As shown in Fig.~\ref{fig:fourier-mask-ratio}, high ratio fixed masking and variable ratio masking effectively amplifies the high frequency information. But interestingly, the experimental results suggest that a high ratio may hinder learning, as performance actually deteriorated, while the use of variable ratio masking resulted in a significant improvement. The low ratio did not lead to any dramatic changes in performance. Table~\ref{tab:ablation-mask} shows that masking with a modest level of \(r_{max}\) proved beneficial for performance by imparting inductive bias without compromising depth information.

\setlength{\tabcolsep}{3pt}
\renewcommand{\arraystretch}{0.7} 
\begin{table}
\caption{\textbf{Ablation study on individual components}: We compared the L1 error performance between two versions: without and with fine-tuning. For the version without fine-tuning, we evaluate on the entire training set, and for the version with fine-tuning, we use the validation set. }
\setlength{\tabcolsep}{2\tabcolsep}
  \centering
  \fontsize{7.5pt}{10.5pt}\selectfont
  \begin{tabular}{cccc|cc}
    \toprule
    Disparity & \multirow{2}{*}{Masking} & Reconstruction & Consistency &  \multicolumn{2}{c}{KITTI 15}\\
    loss& &loss&loss& w/o & w/\\
    \midrule
    \checkmark & & & &1.756&0.550\\
     \checkmark & \checkmark & \checkmark & &1.323&0.538\\
     \checkmark & \checkmark & \checkmark & \checkmark &\textbf{1.179}&\textbf{0.526}\\
    \bottomrule
  \end{tabular}
  \label{tab:ablation_checkmarks}
\end{table}

\setlength{\tabcolsep}{3pt}
\renewcommand{\arraystretch}{0.75} 
\begin{table}[t]
\caption{\textbf{Ablation study with variable mask ratio parameters}: We evaluated each validation set using the bad@1.0\((\)\%\()\)$\downarrow$ metric.}
\centering
\fontsize{8pt}{10.5pt}\selectfont
\begin{tabular}{c c c c c c}
    \toprule
\multirow{4}{*}{Dataset} & \(\mu=0.5\) & \(\mu=0.5\)& \(\mu=0.5\)&\(\mu=0.25\)&\(\mu=0.25\)\\
&\(\sigma=0.1\)&\(\sigma=0.25\)&\(\sigma=0.5\)&\(\sigma=0.5\)&\(\sigma=1.0\)\\

& \(r_{\text{max}}=0.9\)& \(r_{\text{max}}=0.9\)& \(r_{\text{max}}=0.9\)& \(r_{\text{max}}=0.5\) & \(r_{\text{max}}=0.5\)\\
&\(r_{\text{min}}=0.1\) &\(r_{\text{min}}=0.1\)&\(r_{\text{min}}=0.1\)&\(r_{\text{min}}=0.0\) &\(r_{\text{min}}=0.0\)\\ 
    \midrule
SF (c) & 5.1& 5.2 & 5.6 & \underline{4.5} & \textbf{4.2}\\ 
SF (f) &  5.4& 5.4 & 6.0 & \underline{4.5} & \textbf{4.3}\\ 
ETH &  0.95& 3.27 & 2.64 & \textbf{0.26} & \underline{0.27}\\ 
MB & 20.5& 17.9& 22.0& \underline{12.3} & \textbf{11.3}\\ 
  \bottomrule
\end{tabular}
\label{tab:ablation-mask}
\end{table}

\section{Conclusion}
We proposed UniTT-Stereo to maximize the strengths of Transformer-based architecture, which have traditionally lagged behind in stereo matching task. We enhance performance in a simple yet effective manner by employing reconstruction-and-prediction strategy and a combination of losses specifically designed to learn stereo information. Our approach achieves state-of-the-art performance on prominent stereo datasets and demonstrates strong zero-shot generalization capabilities. Throughout this process, we have analyzed the specific advantages our approach brings to stereo depth estimation.


\bigskip

\bibliography{aaai25}

\end{document}